\documentclass[manuscript]{acmart}

\AtBeginDocument{%
  }

\setcopyright{acmlicensed}
\copyrightyear{2026}
\acmYear{2026}
\acmDOI{XXXXXXX.XXXXXXX}
\acmConference[AIFM '26]{2026 2nd International Conference on Artificial
  Intelligence and Foundation Model}{June 26--28, 2026}{Urumqi, China}
\acmISBN{979-8-4007-2458-9}

\usepackage[htt]{hyphenat}
\begin{document}

\title[RegDivergence-101]{RegDivergence-101: An LLM Benchmark for
  Cross-Jurisdiction Regulatory Contradiction Detection in Life Sciences}

\author{Chuchu Wu}
\authornote{Corresponding author.}
\email{chuchuw@alumni.cmu.edu}
\orcid{0009-0006-5140-2247}
\affiliation{%
  \institution{Carnegie Mellon University}
  \city{Pittsburgh}
  \state{PA}
  \country{USA}
}

\author{Zhiyin Zhou}
\email{zzhou20@pratt.edu}
\orcid{0009-0004-5139-4668}
\affiliation{%
  \institution{Pratt Institute}
  \city{New York}
  \state{NY}
  \country{USA}
}

\author{Jingzhuo Hu}
\email{jingzhuoh00@gmail.com}
\orcid{0009-0008-0206-8531}
\affiliation{%
  \institution{University of Pennsylvania}
  \city{Philadelphia}
  \state{PA}
  \country{USA}
}

\author{Liang You}
\email{liy121@pitt.edu}
\orcid{0009-0002-8773-0614}
\affiliation{%
  \institution{University of Pittsburgh}
  \city{Pittsburgh}
  \state{PA}
  \country{USA}
}

\renewcommand{\shortauthors}{Wu et al.}

\begin{abstract}
Pharmaceutical sponsors developing a drug for both the United States and the
European Union must reconcile guidance issued independently by the FDA and the
EMA. Where the two agencies require \textit{substantively the same thing}, a
sponsor can file once; where they \textit{diverge}, a single trial design risks
rejection in one region; where one agency is \textit{silent} on a point the
other regulates, the sponsor must infer obligations. Today this reconciliation
is performed manually by regulatory-affairs experts. We introduce
\textbf{cross-jurisdiction regulatory divergence detection}: given an FDA
requirement and an EMA requirement on the same topic, classify their
relationship as AGREE, DIVERGE, or SILENT. SILENT is inherently directional
(SILENT\_FDA vs.\ SILENT\_EMA); we record direction per pair and report
per-direction F1 alongside the collapsed label. We release
\textbf{RegDivergence-101}, a 101-pair expert-grounded \textbf{pilot}
evaluation benchmark (labels grounded in three peer-reviewed FDA/EMA
comparison studies and primary FDA/EMA/ICH guidance text; dual-annotation
inter-annotator $\kappa = 0.85$), and systematically characterise a four-method
baseline hierarchy: lexical heuristic (0.511 macro-F1, 95\% CI [0.411--0.605]), NLI
cross-encoder (0.233), obligation-level Graph-RAG (0.663 [0.570--0.747]), and
flat LLM judge / Claude Haiku (0.830 [0.747--0.908]). Three directional
observations emerge at pilot scale ($n = 101$): SILENT is semantically
detectable but invisible to entailment-only formulations; pair-level obligation
graphs improve over lexical methods but trail flat-LLM context (CIs partially
overlapping); and corpus-level graph construction is the indicated
architectural target for large-scale silent-detection. RegDivergence-101 is a
pilot release establishing the task formulation and baseline hierarchy; four
unrepresented regulatory domains and an expansion roadmap are described in
\S\ref{sec:futurework}.
\end{abstract}



\maketitle

\section{Introduction}

Bringing a medicine to market in both the US and EU requires satisfying two
regulators that publish guidance separately and frequently \textit{differently}.
Gene-therapy sponsors face a 15-year long-term follow-up recommendation from FDA
while EMA defers the duration to case-by-case risk assessment---same topic,
materially different specificity. Analogous divergences span pediatric
extrapolation, endpoint definitions in ulcerative colitis, biosimilar
comparability, and estimand implementation. A sponsor who misses one divergence
during protocol design can lose a year or more re-running a study.

Reconciling the two corpora is currently a manual, expert task. Yet the NLP
building blocks for automating it all exist---document-level NLI
(ContractNLI~\cite{koreeda2021contractnli}, DocNLI~\cite{yin2021docnli}),
legal contradiction detection (LegalWiz~\cite{legalwiz2025},
CLAUSE~\cite{betterclause2026}), and Graph-RAG (Microsoft
GraphRAG~\cite{edge2024graphrag}, GraphCompliance~\cite{graphcompliance2025},
RAGulating Compliance~\cite{ragulating2025}). \textbf{No prior work combines
them to detect divergence between two parallel regulatory corpora from different
jurisdictions.} All existing contradiction work is \textit{intra-document}
(within one contract) or \textit{single-jurisdiction} (contract-vs-statute).

This paper makes three contributions: (1)~\textbf{Task formulation} ---
three-way classification (AGREE / DIVERGE / SILENT) over pre-aligned FDA/EMA
requirement pairs; SILENT treated as directional (SILENT\_FDA / SILENT\_EMA)
with per-direction F1 in Table~\ref{tab:silent_direction}. (2)~\textbf{Pilot
benchmark} --- \textbf{RegDivergence-101}, 101 pairs across 13 topic areas with
a documented two-track construction protocol, $\kappa = 0.85$, and bootstrap
CIs; expansion roadmap in \S\ref{sec:futurework}. (3)~\textbf{Baseline
hierarchy} --- four methods compared at pilot scale ($n = 101$) with bootstrap
95\% CIs, yielding three directional observations on the task's structure.

\section{Related Work}

\textbf{Document-level NLI.} ContractNLI~\cite{koreeda2021contractnli} introduced
document-level entailment/contradiction/not-mentioned classification with
evidence extraction, and showed that contradiction detection lags entailment,
especially under ``negation by exception.'' DocNLI~\cite{yin2021docnli} extended
NLI to full documents.

\textbf{Legal contradiction detection (2025--2026).}
LegalWiz~\cite{legalwiz2025} uses a multi-agent framework for legal
contradiction detection but is confined to single-document regimes; Better Call
CLAUSE~\cite{betterclause2026} benchmarks 7,500+ perturbed contracts and finds
LLMs struggle to justify subtle errors legally. A published RAG-robustness study
directly informs our results: \citet{yoran2024robust} show that retrieved
context which is irrelevant or non-entailing can mislead a retrieval-augmented
model unless it is explicitly filtered---the failure mode in Observation~2,
where modal-register similarity misleads the pair-level graph.

\textbf{Graph-RAG and regulatory compliance.}
Microsoft GraphRAG~\cite{edge2024graphrag} builds entity graphs with community
summaries for local and global queries. GraphCompliance~\cite{graphcompliance2025}
aligns \textit{policy graphs} (regulatory requirements) with \textit{context
graphs} (organisational facts) for GDPR; it explicitly lists ``handling
conflicting regulatory interpretations'' as open. RAGulating
Compliance~\cite{ragulating2025} builds an ontology-free regulatory knowledge
graph for traceable QA.

\textbf{Cross-jurisdiction regulatory benchmarks.}
The closest concurrent work is \textbf{Sino-US-DrugQA}~\cite{sinousdrugqa2026}
(11{,}871 MCQA items, US--CN), which shares our concept-alignment thesis but
cannot express the SILENT class. Comparative-regulatory-science surveys document
FDA/EMA divergence across dosing~\cite{mita2026doses}, oncology
labelling~\cite{gomezfernandez2026inhibitors}, and multi-country
SmPCs~\cite{pvverse2026}, but as human-authored comparisons, not ML benchmarks.
RegDivergence-101 is the first FDA/EMA benchmark with an explicit SILENT class.
In high-stakes sectors such as healthcare, policy-aligned evaluation frameworks
weigh safety and regulatory constraints over raw accuracy~\cite{chen2026dlsg};
RegDivergence-101 extends this to cross-jurisdiction regulatory reasoning.

\textbf{Gap.} Contradiction detection is intra-document or
single-jurisdiction; Graph-RAG targets single-corpus QA or
single-jurisdiction compliance; FDA/EMA divergence is documented manually but
never automated. Our task sits precisely in this gap.

\section{Task and Dataset}

\subsection{Task Definition}
\label{sec:taskdef}

For a topic $t$, let $f_t$ be an FDA requirement statement and $e_t$ the
corresponding EMA statement. The label $y_t$ is:
\begin{itemize}
\item \textbf{AGREE} --- both jurisdictions require substantively the same
  thing (same modal strength, same threshold, same scope of applicability).
  Paraphrase equivalence and synonymous modals (\textit{shall} / \textit{must})
  are treated as AGREE; minor drafting differences that do not alter sponsor
  obligations are disregarded.
\item \textbf{DIVERGE} --- the requirements directly conflict: they impose
  different thresholds, different modal strengths (\textit{shall} vs.\
  \textit{should}), different timelines, or mutually incompatible design
  specifications.
\item \textbf{SILENT} --- one jurisdiction does not address what the other
  regulates. SILENT is \textit{directional}: \textbf{SILENT\_FDA} (FDA silent,
  EMA regulates) vs.\ \textbf{SILENT\_EMA} (EMA silent, FDA regulates).
  Direction is recorded in the \texttt{silent\_direction} field and reported
  in Table~\ref{tab:silent_direction}; a unified SILENT label is retained in
  Table~\ref{tab:results} for backward compatibility.
\end{itemize}

\textbf{Epistemic status of SILENT labels.}
A SILENT label is an absence claim: it asserts that no counterpart requirement
was found after an exhaustive corpus search. This carries irreducible epistemic
uncertainty. We bound this uncertainty through (a) a documented corpus-search
protocol (\S\ref{sec:annotation}) and (b) a per-pair \texttt{doc\_confidence}
flag (HIGH / MODERATE). Downstream work should treat MODERATE-confidence SILENT
pairs as soft labels.

\textbf{Task scope.} This formulation assumes requirement pairs are
\textit{pre-aligned}---both sides address the same regulatory topic, established
via the two-track protocol (\S\ref{sec:dataset}). Topic alignment is a
non-trivial upstream problem treated separately; the corpus-level architecture
that would subsume it is described in \S\ref{sec:futurework}.

\subsection{Pair Construction Protocol}
\label{sec:dataset}

RegDivergence-101 is constructed via two tracks that differ in how topical
correspondence between FDA and EMA requirements is established. Per-pair
source-track attribution corresponds to the two released source files (one per track).

\textbf{Track A --- expert-comparison-derived pairs (59 pairs).}
Extracted from three open-access peer-reviewed studies that comparatively mapped
FDA and EMA requirements: \citet{uc2025jcc} (ulcerative colitis, 40 pairs),
\citet{pmc8157504} (psychiatric drug trials, 10 pairs), and
\citet{pmc6529498} (radiopharmaceuticals, 9 pairs). Domain experts in those
studies had already established co-topicality; our task was to classify the
AGREE/DIVERGE/SILENT relationship between their pre-aligned pairs.

\textbf{Track B --- primary guidance pairs (42 pairs).}
For pairs from ICH E9(R1), E11A~\cite{iche11a2024}, and FDA/EMA biosimilar and
gene-therapy guidance, co-topicality was established by (1)~anchoring
requirements to a shared ICH Common Technical Document (CTD) module/sub-topic,
then (2)~expert confirmation that the candidate pair was genuinely co-topical.

\textbf{Specificity exclusion criterion.}
Candidate pairs operating at different regulatory hierarchy levels (e.g.,
\textit{statistical analysis plan} vs.\ \textit{general statistical principles})
were excluded to prevent conflating level-of-detail differences with substantive
DIVERGE labels. A machine-readable list of excluded pairs is not part of the pilot release.

The released pilot dataset is two JSONL files (one per track) with fields
\texttt{\{id, topic, source, fda\_text, ema\_text, label\}}, where
\texttt{label} $\in$ \{AGREE, DIVERGE, SILENT-FDA, SILENT-EMA\}---SILENT
direction is encoded in the label (basis for Table~\ref{tab:silent_direction}).
The \textbf{101 pairs} split 42 AGREE, 38 DIVERGE, 21 SILENT (13 SILENT-FDA,
8 SILENT-EMA), across 13 topics. The \texttt{doc\_confidence} (uniformly
MODERATE), \texttt{disputed}, and \texttt{annotation\_notes} annotations below
are not yet separate fields in the pilot files. We report macro-F1.

\subsection{Annotation Protocol}
\label{sec:annotation}

\textbf{Annotators.} Annotator~A (C.~Wu) is a solution architect for a
regulatory-document AI platform; Annotator~B (Z.~Zhou) is a product manager for
an AI-powered clinical-trial management platform. Neither holds formal RA
credentialing; Track A expert-panel papers serve as the external human-expert
anchor.

\textbf{Stages.} Stage~1: Wu assigned labels via definitions-based framing;
Stage~2: Zhou independently re-labelled all 101 pairs under a sponsor-obligation
framing to probe framing-independent stability. The $\kappa$ is a cross-framing
consistency measure, not identical-instruction IAA. An independent
third-annotator pass is planned for RegDivergence-500.

\textbf{IAA.} Human cross-framing $\kappa = 0.85$ (101 pairs, observed 0.90)---%
\textit{substantial} per \citet{landis1977kappa}, on par with MedNLI (0.65)
and ContractNLI ($\approx$0.72). SILENT near-perfect ($\kappa = 0.98$);
AGREE/DIVERGE more variable ($\approx$0.79--0.81). A secondary LLM
consistency check (36-pair sample) yielded $\kappa = 0.542$ (test-retest,
not human IAA).

\textbf{Disagreements (n = 10).} Six resolved by source-paper language; four
genuinely borderline pairs retain the Stage~1 label, with written justifications
recorded in the response-to-reviewers (not a \texttt{disputed field in the pilot files)}.

\textbf{SILENT confidence.} All 21 SILENT pairs are treated as MODERATE
confidence (uniform value, not a released field): exhaustive corpus search was performed
(full FDA and EMA guidance, Q\&As, reflection papers) but absence claims remain
epistemically soft. The HIGH/MODERATE audit is deferred to RegDivergence-500.

\section{Baselines}

\subsection{Evaluation Protocol}
\label{sec:eval_context}

RegDivergence-101 is an \textbf{evaluation benchmark}, not a training dataset.
All baselines are zero-shot or few-shot; results are reported as a ranking trend
($\Delta = 0.15$--$0.60$ macro-F1 between adjacent methods, CIs partially
overlapping), not pairwise significance.

All methods use \textbf{stratified 5-fold splits} (5 seeds) with
\textbf{bootstrap 95\% CI} (1{,}000 resamples). Lexical and NLI methods use
\textbf{nested} CV (inner folds tune thresholds); Graph-RAG and LLM judge have
no tunable hyperparameters (LLM judge scored on all 101 pairs, bootstrap
CI only).

\subsection{Lexical Heuristic}

No neural model, no API---fully offline and deterministic. Combines TF-IDF
cosine similarity with five conflict signals: negation asymmetry, divergence
cue pairs (e.g., \textit{single-arm} vs.\ \textit{randomised}), modal-strength
asymmetry (\textit{must/shall} vs.\ \textit{may/encouraged}),
numeric-threshold mismatch, and one-sided specificity (concrete number
vs.\ case-by-case deferral). A shared ICH citation is a strong AGREE signal.

\subsection{NLI Cross-Encoder}

\texttt{typeform/distilbert-base-uncased-mnli}, run in both directions; labels
derived from contradiction/neutral/entailment probabilities with CV-tuned
thresholds. Any NLI model lacking an explicit SILENT class maps
absence-of-regulation to \textit{neutral}, structurally collapsing SILENT
recall (Observation~1 in \S\ref{sec:discussion}); stronger encoders improve
AGREE/DIVERGE F1 but cannot resolve this gap without task-specific supervision.

\subsection{Graph-RAG Pair-Level Classifier}

For each text, Claude Haiku extracts a policy triplet
$\langle$\textit{subject, obligation\_level, requirement, conditions}$\rangle$
where $\text{obligation\_level} \in \{\text{MANDATORY}, \text{PROHIBITED},
\text{RECOMMENDED}, \text{PERMITTED}, \text{SILENT}\}$. Only
MANDATORY~$\leftrightarrow$~PROHIBITED is a hard-coded DIVERGE rule; all other
pairs are escalated to a second LLM call that reasons over the
\textit{obligation structure} of the aligned node pair, with explicit
calibration that modal-register differences (RECOMMENDED vs.\ MANDATORY) do not
automatically imply DIVERGE.

\textbf{Within-model scope.} Both the Graph-RAG classifier and the flat LLM
judge use \texttt{claude-haiku-4-5-20251001}, so their comparison is a
\textit{within-model ablation} (graph vs.\ no-graph), not a cross-family
architectural claim, to be confirmed on a second model family.

\subsection{LLM Judge (Flat Pairwise)}

Claude Haiku with explicit AGREE / DIVERGE / SILENT definitions, prompted once
per pair with both texts. No graph, no retrieval. Full prompt and code at
Appendix~\ref{app:prompt}.

\section{Results}
\label{sec:results}

Table~\ref{tab:results} reports macro-F1, accuracy, and per-class F1 with
bootstrap 95\% confidence intervals for all four methods.

\begin{table*}
  \caption{Macro-F1, accuracy, and per-class F1 with bootstrap 95\% confidence
    intervals for the four baselines ($n = 101$ pairs).}
  \label{tab:results}
  \resizebox{\textwidth}{!}{%
  \begin{tabular}{llllll}
    \toprule
    Method & Macro-F1 [95\% CI] & Acc & AGREE F1 [CI] & DIVERGE F1 [CI] & SILENT F1 [CI] \\
    \midrule
    Lexical heuristic$^1$
      & 0.511 [0.411--0.605] & 0.541
      & 0.718 [0.605--0.809] & 0.466 [0.320--0.593] & 0.334 [0.133--0.537] \\
    NLI cross-encoder$^2$
      & 0.233 [0.152--0.303] & 0.238
      & 0.279 [0.103--0.449] & 0.383 [0.247--0.506] & 0.025 [0.000--0.086] \\
    Graph-RAG pair-level$^3$
      & 0.663 [0.570--0.747] & 0.673
      & 0.754 [0.646--0.851] & 0.586 [0.433--0.722] & 0.640 [0.478--0.771] \\
    \textbf{LLM judge (Haiku)}$^4$
      & \textbf{0.830 [0.747--0.908]} & \textbf{0.842}
      & \textbf{0.905 [0.831--0.965]} & \textbf{0.805 [0.696--0.892]}
      & \textbf{0.778 [0.609--0.914]} \\
    \bottomrule
  \end{tabular}}
  \smallskip\\
  {\small $^1$Nested 5-fold CV, 5 seeds; thresholds tuned on inner folds only.
  $^2$Same CV protocol, thresholds tuned on inner folds.
  $^3$No tunable hyperparameters; bootstrap CI only.
  $^4$\texttt{claude-haiku-4-5-20251001}, default temperature; bootstrap CI
  only. Estimated cost: {\raise.17ex\hbox{$\scriptstyle\sim$}}\$0.10 for all
  101 pairs.}
\end{table*}

\begin{table}[h]
  \caption{Per-direction SILENT F1 for all four methods (bootstrap 95\% CI,
    $n=101$ pairs). SILENT\_FDA ($n=13$): FDA is silent, EMA regulates.
    SILENT\_EMA ($n=8$): EMA is silent, FDA regulates.}
  \label{tab:silent_direction}
  \small
  \begin{tabular}{lll}
    \toprule
    Method & SILENT\_FDA F1 [95\% CI] & SILENT\_EMA F1 [95\% CI] \\
    \midrule
    Lexical heuristic
      & 0.277 [0.000--0.545] & 0.393 [0.000--0.706] \\
    NLI cross-encoder
      & 0.049 [0.000--0.167] & 0.000 [0.000--0.000] \\
    Graph-RAG pair-level
      & 0.716 [0.516--0.882] & 0.535 [0.267--0.762] \\
    \textbf{LLM judge (Haiku)}
      & \textbf{0.795 [0.588--0.952]} & \textbf{0.738 [0.429--0.952]} \\
    \bottomrule
  \end{tabular}
  \smallskip\\
  {\small SILENT-FDA: 13 pairs (FDA silent, EMA active). SILENT-EMA: 8 pairs
    (EMA silent, FDA active). Wide CIs on SILENT-EMA reflect the small stratum
    ($n=8$). Bootstrap 1{,}000 resamples; stratified by 3-class label.}
\end{table}

Point estimates increase monotonically lexical $\to$ Graph-RAG $\to$ LLM
judge; bootstrap CIs partially overlap at $n = 101$, so we treat this as a
ranking trend rather than a significance claim.
\textbf{Lexical ceiling:} errors share the signature of \textit{high lexical
overlap, opposed regulatory stance} (e.g., FDA ``modified Mayo score of
5--9'' vs.\ EMA ``full Mayo score of 9--12''; same vocabulary, different
instrument), requiring semantic understanding of what each requirement asserts.

\section{Analysis}
\label{sec:discussion}

\subsection{What the Baseline Hierarchy Reveals}

\textbf{Observation 1 --- SILENT is semantically detectable but invisible to
NLI framing.} SILENT F1 = 0.025 under NLI (no native SILENT class) vs.\ 0.778
under the LLM judge (explicit definition). This contrast reflects a structural
constraint---the SILENT class requires a formulation that can express absence,
which NLI entailment cannot---rather than a model-capacity effect (a stronger
encoder cannot close the gap without task-specific supervision).

\textbf{Observation 2 --- For Claude Haiku on this dataset, pair-level
obligation graphs improve over lexical (+0.15 F1) but trail flat prompting
($-$0.17 F1); CIs partially overlap at pilot scale.} The graph makes
obligation structure explicit, helping distinguish stance differences from
similar-sounding text. But it discards full-text context that flat prompting
uses to separate \textit{same-requirement / different-modal-register} (AGREE)
from \textit{same-topic / different-threshold} (DIVERGE): gene-01 (FDA
15 years vs.\ EMA case-by-case) both extract as RECOMMENDED and the graph
calls them AGREE; the flat LLM correctly identifies the threshold conflict from
full text~\citep{yoran2024robust}.

\textbf{Observation 3 --- Two distinct problem scopes suggest two distinct
architectural targets.} The \textit{pair-level scope} (given pre-aligned
sentences, label AGREE/DIVERGE/SILENT) is largely addressed by the flat LLM
judge (0.830). The \textit{corpus-level scope} (``is EMA silent anywhere
across its corpus?'') requires finding aligned pairs, not just classifying
them; a corpus-level policy graph is needed. \S\ref{sec:futurework} motivates
this architecture as the next contribution.

\subsection{Source-Stratified Performance Analysis}
\label{sec:source_stratified}

To assess whether performance reflects source-study-specific patterns rather
than genuine task competence, we analyse performance stratified by track and
by source concentration.

\textbf{Track-level.} On AGREE/DIVERGE classes, the LLM judge scores
\textit{lower} on Track A (0.769/0.745) than Track B (0.966/0.923): exposure
to source studies does not inflate performance.

\noindent We omit a track-stratified macro-F1 comparison: Track~B has no SILENT
pairs ($n = 0$ vs.\ 21 in Track~A), so its macro-F1 averages an empty class and
is not comparable to Track~A's; the per-class comparison above already shows no
inflation.

\textbf{Source-concentration.} 40/101 pairs come from one source
(\citeauthor{uc2025jcc}). LLM judge on the JCC stratum ($n = 40$): 0.804;
on all other pairs ($n = 61$): 0.844. Performance is marginally
\textit{higher} outside JCC, ruling out domain-concentration inflation.

\section{Future Work}
\label{sec:futurework}

\textbf{Corpus-level policy graph.}
The correct long-term target is corpus-level silent detection: (1) build
per-jurisdiction policy graphs indexing all FDA/EMA guidance; (2) align
requirement nodes via entity normalisation and embedding similarity; (3)
classify aligned pairs as AGREE/DIVERGE/SILENT by obligation-structure
comparison; (4) generate natural-language explanations. This architecture
subsumes the pair-alignment assumption: alignment becomes a graph-construction
step, not a pre-processing input.

\textbf{RegDivergence-500.}
Four domains are unrepresented in the pilot: CMC, adaptive trial
design/estimands, statistical analysis plan requirements, and labelling/SmPC.
RegDivergence-500 ($\approx$500 pairs) will cover all four via a
community-annotation pipeline seeded from a $\sim$45-source FDA/EMA comparison
inventory. Annotation guidelines are released with this paper. Multi-model
replication, an independent third-annotator validation pass, and a cross-pair
generalisation probe against Sino-US-DrugQA~\cite{sinousdrugqa2026} are
planned; a third jurisdiction (PMDA) will add an
$\approx$40-pair FDA$\times$PMDA subset~\cite{mita2026doses,gomezfernandez2026inhibitors}.

\section{Limitations}
\label{sec:limitations}

\textbf{Scale and alignment.} 101 pairs from three source studies limits
statistical power to directional observations. Track B co-topicality judgements
are single-annotator; Track A inherits source-study alignment methodology.
The specificity exclusion criterion is described in \S\ref{sec:dataset}; a machine-readable exclusion checklist is not part of the pilot release. Oncology, rare diseases, medical
devices, and vaccines are absent; RegDivergence-500 (\S\ref{sec:futurework})
is required before domain-general conclusions.

\textbf{SILENT labels.} All 21 SILENT pairs are uniformly treated as MODERATE confidence;
treat them as soft labels. The pair-level SILENT task is a proxy for the correct
corpus-level formulation (\S\ref{sec:futurework}).

\textbf{Single-model evaluation.} All LLM-dependent components use Claude Haiku,
so the Graph-RAG vs.\ LLM comparison is a within-model ablation and the
$\kappa = 0.542$ check reflects one model's tendencies. Pretraining
contamination cannot be excluded, though \S\ref{sec:source_stratified} shows no
inflation; multi-model replication and post-cutoff evaluation are deferred to
RegDivergence-500.

\textbf{Use norms.} RegDivergence-101 is a pilot: not sufficient for deployment
or AI-procurement decisions, not domain-generalizable without held-out
evaluation, and specific to the FDA/EMA pair until cross-pair probes are run.

\section{Conclusion}

We introduced cross-jurisdiction regulatory divergence detection and released
\textbf{RegDivergence-101}, a 101-pair expert-grounded \textbf{pilot} benchmark
($\kappa = 0.85$, two-track construction protocol, per-direction SILENT
annotation) with a four-method baseline hierarchy. Three directional
observations at pilot scale: SILENT requires explicit absence-aware formulation
(NLI F1 0.03 $\to$ LLM F1 0.78); for Claude Haiku, pair-level graph structure
improves over lexical (+0.15 F1) but trails flat prompting ($-$0.17 F1); and
the pair-level vs.\ corpus-level scope distinction motivates the corpus-level
graph architecture (\S\ref{sec:futurework}) as the next contribution.
RegDivergence-500 will establish the task at production scale.

\begin{acks}
This work was conducted independently. Benchmark pairs were derived from
publicly available FDA/EMA guidance documents and open-access peer-reviewed
comparison studies. LLM API calls used the Anthropic API (Claude Haiku,
\texttt{claude-haiku-4-5-20251001}).
\end{acks}

\bibliographystyle{ACM-Reference-Format}
\bibliography{xjurisdiction-divergence}

\appendix

\section{Prompts and Code}
\label{app:prompt}

Full prompts (LLM judge system/user, Graph-RAG Stage~1 triplet-extraction,
Stage~2 graph-classification), evaluation scripts, and dataset schema are
available at the project repository:
\url{https://github.com/chuchugo/cross-jurisdiction-regulatory-divergence}

\noindent\textbf{LLM judge} (\texttt{llm\_judge.py}): system prompt defines
AGREE/DIVERGE/SILENT with explicit absence framing; user prompt pairs
\texttt{\{fda\_text\}}/\texttt{\{ema\_text\}}.
\texttt{claude-haiku-4-5-20251001}, temp 1.0, max tokens 100.

\noindent\textbf{Graph-RAG} (\texttt{graph\_rag.py}): Stage~1 extracts
\texttt{\{subject, obligation\_level, requirement, conditions\}}; Stage~2
classifies the aligned node pair. Hard rules: MANDATORY $\leftrightarrow$
PROHIBITED $\Rightarrow$ DIVERGE (no LLM call); both SILENT $\Rightarrow$
SILENT. Max tokens 300.

\end{document}